\documentclass{article}

\PassOptionsToPackage{numbers}{natbib}
\usepackage[preprint]{neurips_2019}

\usepackage[utf8]{inputenc} 
\usepackage[T1]{fontenc}    
\usepackage{hyperref}       
\usepackage{cleveref}
\usepackage{url}            
\usepackage{booktabs}       
\usepackage{amsfonts}       
\usepackage{nicefrac}       
\usepackage{microtype}      
\usepackage{graphicx}
\usepackage{bm}
\usepackage{amsmath} 
\usepackage{amssymb}
\usepackage{amsthm}
\usepackage{amsxtra}
\newtheorem{theorem}{Theorem}[section]
\newtheorem{definition}[theorem]{Definition}
\usepackage{soul}               
\usepackage[color=green!40]{todonotes}

\title{On the Transformation of Latent Space \\
in Autoencoders}

\author{%
  Jaehoon Cha \\
  \texttt{jaehoon.cha@xjtlu.edu.cn} \\
  \And
  Kyeong Soo Kim \\
  \texttt{kyeongsoo.kim@xjtlu.edu.cn} \\
  \And
  Sanghyuk Lee \\
  \texttt{sanghyuk.lee@xjtlu.edu.cn} \\
  \AND
  \\
  Department of Electrical and Electronic Engineering \\
  Xi'an Jiaotong-Liverpool University \\
  Suzhou, 215123, P. R. China \\
}%

\graphicspath{{./plots/}, {./figures/}}

\begin{document}

\maketitle

\begin{abstract}
  Noting the importance of the latent variables in inference and learning, we
  propose a novel framework for autoencoders based on the homeomorphic
  transformation of latent variables --- which could reduce the distance between
  vectors in the transformed space, while preserving the topological properties
  of the original space --- and investigate the effect of the latent space
  transformation on learning generative models and denoising corrupted data. The
  experimental results demonstrate that our generative and denoising models
  based on the proposed framework can provide better performance than
  conventional variational and denoising autoencoders due to the transformation,
  where we evaluate the performance of generative and denoising models in terms
  of the Hausdorff distance between the sets of training and processed---i.e.,
  either generated or denoised---images, which can objectively measure their
  differences, as well as through direct comparison of the visual
  characteristics of the processed images.
\end{abstract}

\section{Introduction}
\label{sec:intro}
Data compression/restoration and generating new data based on the learned
distribution from a training dataset have been extensively studied in the
context of machine learning, especially with artificial neural networks. In
their early stage, the wake-sleep algorithm was used to produce a good density
estimator by training a stack of layers so that each of the layers can correctly
represent activities above and below it \cite{hinton1995wake}.

Recently, Autoencoders (AE) have been gaining huge attention from researchers
not only for data compression/restoration but also as generative models. AE is
originally studied to extract salient features through its bottleneck structure
which reduces the dimensionality of the input data \cite{hinton2006reducing}. AE
is also studied as efficient generative models
\cite{vincent2008extracting,vincent2010stacked,makhzani2015adversarial,sohn2015learning,maaloe2016auxiliary,creswell2018denoising}. In
particular, Variational Autoencoder (VAE) is introduced as a stochastic
variational inference and learning algorithm \cite{kingma2013auto}. The encoder
network of VAE approximates the posterior distribution given the input data and
infers good values of latent variables. Then, the decoder network generates a
distribution of input data over the latent variables. Because VAE takes latent
variables during the generation phase, we began to realize the importance of the
latent space and regularizers and investigate ways on how to explore the latent
space during training and generating processes.

Note that the training objectives and the use of the reparameterization trick
with Gaussian latent variables in conventional VAE may result in a regularizer
with poor inference quality and thereby provide models which are not able to
properly capture the dependencies in the original data due to the assumption of
independent latent variables
\cite{maaloe2016auxiliary,zhao2017infovae,tolstikhin2017wasserstein}. In this
paper, therefore, we propose a novel framework of Latent space Transformation in
Autoencoder (LTAE) based on the idea of mapping the latent space through
transformation technique, which requires two new steps compared to the
conventional AE: First, we reduce the distance between any two vectors in the
latent space through the proposed transformation technique, which acts as a
regularizer. Second, we explore the area of the latent space which does not
correspond to any input vector through adding noise to the output from the
encoder network in order to deal with unseen data. The LTAE framework also
provides a better connection from inputs to latent variables to outputs by
eliminating the reparameterization trick used in conventional VAE.

The advantages of the proposed LTAE framework are two-fold: First, this
framework is so flexible that it can be applicable to both generative and
denoising models. Second, the framework could improve the performance of the
resulting models compared to conventional AEs.

Note that for reconstruction applications, the LTAE framework can be applied to
Denoising Autoencoder (DAE), which was invented to extract more useful features
by introducing a new training principle of denoising partially corrupted input
data \cite{vincent2008extracting,vincent2010stacked}. The introduced noise
enables DAE to find useful features in a more robust way and results in good
performance when reconstructing corrupted data. Denoising Latent space
Transformation in Autoencoder (DLTAE)---i.e., DAE based on the LTAE
framework---introduces noise at two different spaces in training, i.e., the
input space and the latent space, to further enhance the robustness of a
resulting model. Due to the transformation in DLTAE, it is also capable of
generating data by taking variables in the transformed latent space with the
decoder network. The generated images are clearer than those by VAE and LTAE,
because DLTAE can capture more salient features by the noise introduced at two
different spaces.

We also propose the use of the Hausdorff distance \cite{munkres1975topology} as
an objective measure of the performance of generative and denoising models,
which is frequently used in computer vision and pattern recognition to measure
the extent to which each point of a model set lies near some point of an image
set and vice versa and thereby provide a degree of resemblance between the two
\cite{huttenlocher93:_compar_hausd}. Note that, however, we extend the
application of the Hausdorff distance to the measurement of the
\textit{similarity between two sets of images} (i.e., the set of training images
and that of processed images), rather than the similarity between two individual
images/shapes, in this paper.


\section{Preliminaries}
\label{sec:pre}
\subsection{Notations and Basic Definitions}
\label{subsec:nb}
$\mathbb{R}^d$ denotes a $d$-dimensional Euclidean space. Vectors are written in
bold lowercase. If $\bm{x}$ is a vector, then, its $i$th element is denoted by
$x_i$. We use bold uppercase letters for matrices (e.g., $\bm{A}$).
\begin{definition}
  A nonempty set $A$ in a metric space $(X,d)$ is said to be bounded if the
  diameter $\operatorname{diam}(A){<}\infty$, where
  \begin{equation}
    \operatorname{diam}(A) \triangleq \sup_{x, y \in A} d(x, y) .
  \end{equation}
\end{definition}
\begin{theorem}
  A sequence $(\bm{x}_n)$ in a normed space $X$ is convergent if $X$ contains an
  $\bm{x}$ such that
  \begin{equation}
    \lim_{n \rightarrow \infty} \|\bm{x}_n - \bm{x} \| = 0.
  \end{equation}
  Then we write $\bm{x}_n{\rightarrow}\bm{x}$.
\end{theorem}
\begin{theorem}\cite{kreyszig1978introductory}
  Let $B$ be a subset of a metric space $X$ and let $\varepsilon{>}0$ be
  given. A set $M_{\varepsilon}{\subset}X$ is called an $\varepsilon{-}$net for
  $B$ if for every point $z{\in}B$ there is a point of $M_{\varepsilon}$ at a
  distance from $z$ less than $\varepsilon$. The set $B$ is said to be totally
  bounded if for every $\varepsilon > 0$ there is a finite $\varepsilon{-}$net
  $M_{\varepsilon}{\subset}X$ for $B$, where ``finite'' means that
  $M_{\varepsilon}$ is a finite set.
\end{theorem}
\begin{theorem}\cite{gamelin1999introduction}
  A subset $E$ of $\mathbb{R}^n$ is totally bounded if and only if $E$ is
  bounded.
\end{theorem}
\begin{definition}\cite{munkres1975topology}
  Let $X$ and $Y$ be topological spaces and $f: X{\rightarrow}Y$ be a bijection,
  which is a one-to-one (injective) and onto (surjective) mapping. The function
  $f$ is called a homeomorphism if $f$ and the inverse function
  $f^{-1}:Y{\rightarrow}X$ are continuous, and $X$ and $Y$ with a homeomorphism
  are called homeomorphic.
\end{definition}
\begin{definition}\cite{munkres1975topology}
  Let $(X,d)$ be a metric space. If $U{\subset}X$ and $\epsilon{>}0$, let
  $B(U,\epsilon)$ be the $\epsilon{-}$neighborhood of $U$. Let $\mathcal{H}$ be
  the collection of all (nonemepty) closed, bounded subsets of $X$. If
  $U,V{\in}\mathcal{H}$, then the Hausdorff distance is defined by
  \begin{equation}
    D(U,V) = \inf\{\epsilon|U \subset B(V,\epsilon) \, \rm{and} \, V \subset B(U,\epsilon)\}
  \end{equation}
\end{definition}
It is equivalent to
\begin{equation}
  \label{eq:hausdorff}
  D(U,V) = max\{\sup_{u\in U}\inf_{v\in V}d(u,v),\sup_{v\in V}\inf_{u\in U}d(u,v)\}
\end{equation}
Note that a space in this paper refers to a normed space unless stated
otherwise. We use upper case letters to denote spaces (e.g., $X$). Especially,
$X_{in}$ and $X_{out}$ denotes the input space and the output space,
respectively.

\subsection{Problem Statement}
For unsupervised learning, the encoder network reduces the dimension of inputs,
which enables the AE to capture the important features of the original
data. Then the decoder network restores the original data from the compressed
representation. The weights in the AE are updated to closely match the original
data by backpropagation \cite{hinton2006reducing}.

Here we focus on the hidden space between the encoder and the decoder network of
an AE, which we call \textit{a latent space} and denote by $Z$.\footnote{If we
  consider the encoder network as a function, the latent space corresponds to
  the image of the function.} The main goal of this work is to transform vectors
in the latent space to improve the performance of a generative model based on
the decoder network.
In the original AE, a neural network consisting of an encoder network $f$ and a
decoder network $g$ with weights and biases $\phi$ and $\theta$ is trained to
minimize the following loss function:
\begin{equation}
  \label{eq:ae}
  \frac{1}{N}\sum_{\bm{x} \in X_{in}} L\left(\bm{x}, g(f(\bm{x};\phi);\theta)\right), 
\end{equation}
where $f(\bm{x};\phi){\in}Z$, $N$ is the number of input vectors and $L$ is a
loss function which could be either cross-entropy or $L_2$ loss.

Note that there is a set of vectors in the latent space $Z$, which do not
correspond to any input vector. We explore this set by adding noise to the
output from the encoder network in order to make the original AE a generative
model. In the LTAE framework, we introduce a transformation network and a latent
network between the encoder and the decoder network of the original AE to make
it a generative model. The latent network receives the outputs of the encoder
network and injects them to the decoder network. Due to a loss function between
the latent network and the transformation network, vectors are transformed in
the latent network. We denote by $Z_L$ a space of the transformed vectors
through the latent network. By reducing the distances between output vectors in
$Z_L$ without changing their topological properties, the interpolation between
output vectors during the generative phase can be easier and more meaningful. In
this section, a method to make $Z_L$ and to deal with unseen vectors, which are
possibly lie on the sparse spaces on $Z$ or $Z_L$, is described.

\section{Latent Space Transformation in Autoencoder}
\label{sec:ltae}

\subsection{Continuity of the Original Autoencoder}
Let us assume that one layer of a neural network consists of a set of matrix
multiplication, addition, and an activation function. We define a function
$h: X{\rightarrow}Y$, given by
\begin{equation}
  h(\bm{x}) = f(\bm{A}\bm{x} + \bm{b})
\end{equation}
where $\operatorname{dim}(X){=}n$, $\operatorname{dim}(Y){=}r$,
$\bm{A}:\mathbb{R}^n{\rightarrow}\mathbb{R}^r$ is a matrix operator, $\bm{b}$ is
a vector of $r$ components, and $f$ is an activation function such as Softplus,
sigmoid, hyperbolic tangent (tanh), rectified linear unit (ReLU), and leaky
ReLU. Then, $\bm{x}_n{\rightarrow}\bm{x}$ implies
$h(\bm{x}_n){\rightarrow}h(\bm{x})$ from the fact that
\begin{align}
  \| h(\bm{x}_n) - h(\bm{x}) \| &= \| f(\bm{A}\bm{x}_n+\bm{b}) - f(\bm{A}\bm{x}+\bm{b}) \| \\
                                &\leq \|{\bm{A}(\bm{x}_n - \bm{x})}\| \\
                                &\leq \|{\bm{A}}\|\|{\bm{x}_n-\bm{x}}\|,
\end{align}
because matrix multiplication is bounded and all the activation functions
considered satisfy
\begin{equation}
  \label{eq:activation_function}
  \|f(\bm{x}_n)-f(\bm{x})\| \leq \|\bm{x}_n-\bm{x}\|.
\end{equation}

Due to the fact that a composite of continuous functions is continuous and a
network with consecutive layers is a composite of layers, the continuity
preserves through the layers.

Now, we let $f_\phi:X_{in}{\rightarrow}Z$ and $g_\theta:Z{\rightarrow}X_{out}$
be composite functions of hidden layers from (\ref{eq:ae}) where
$f_\phi(\bm{x}){=}f(\bm{x};\phi)$ and $g_\theta(\bm{z}){=}g(\bm{z};\theta)$.  If
$\bm{x}'{=}g_\theta(f_\phi(\bm{x}))$ and $\bm{y}'{=}g_\theta(f_\phi(\bm{y}))$
for any $\bm{x},\bm{y}{\in}X_{in}$, then
\begin{equation}
  \label{eq:bounded}
  \| \bm{x}' - \bm{y}' \|  \leq c_g\| f_\phi(\bm{x}) - f_\phi(\bm{y}) \|  \leq c_f\|
  \bm{x} - \bm{y} \| . 
\end{equation}
where $c_f$ and $c_g$ denote the product of the norms of projection matrices in
the encoder and the decoder networks, respectively. From \eqref{eq:bounded},
therefore, we can expect that, if the distance between latent vectors $\bm{z}$
and $\bm{z'}$, which correspond to an observed vector and an unseen vector in
the input space respectively, is small, the distance between the resulting
outputs from the decoder network in a generative model --- i.e.,
$g_\theta(\bm{z})$ and $g_\theta(\bm{z}')$ --- is small. In fact, this is the
major reason we introduce the latent space transformation in the LTAE framework.

\subsection{Mapping of Unseen Vectors}
\label{subsec:condition}
Let $Z{=}U{\cup}V$, where $U$ is a subset of $Z$ which consists of
$f_\phi(\bm{x})$ for all $\bm{x}{\in}X_{in}$ and $V{=}Z{-}U$. Because $U$ is a
subset of $\mathbb{R}^m$, where $m$ is a dimension of the latent space, and
bounded, it is a totally bounded. Therefore, for every $\varepsilon{>}0 $ there
is a finite $\varepsilon{-}$net $M_\varepsilon$ for $U$. Let
$M_\varepsilon{=}\{\bm{m}_\varepsilon^{(1)}, \bm{m}_\varepsilon^{(2)}, \cdots,
\bm{m}_\varepsilon^{(K)}\}$. Then there is a collection of open balls
$\mathcal{B}{=}\cup_{i=1}^K B_d(\bm{m}^{(i)}, \varepsilon)$ such that
$U{\subset}\mathcal{B}$, where
$B_d(\bm{m}^{(i)},
\varepsilon){\triangleq}\{\bm{z}|d(\bm{m}^{(i)},\bm{z}){<}\varepsilon\}$ and $d$
is a given metric or a metric induced by norm on $Z$. Then, there is
$\bm{m}^{(i)}$ for all $\bm{u}^{(j)}{\in}U$ such that
$B_d(\bm{m}^{(i)}, \varepsilon){\subset}B_d(\bm{u}^{(j)}, 2\varepsilon)$ and it
implies
\begin{equation}
  \label{eq:latent_space_separation}
  Z =  \hat{U} \cup \hat{V},
\end{equation}
where $\hat{U}{=}\cup_{j = 1}^J B_d(\bm{u}^{(j)}, 2\varepsilon)$,
$\hat{V}{=}Z{-}\hat{U}$ and
$K{\leq}J{\leq}N$.

Note that, unlike $U$, $\hat{U}$ in \eqref{eq:latent_space_separation} now
includes latent vectors corresponding to both unseen vectors and observed
vectors in the input space, i.e., $\hat{U}{\supset}U$. $\hat{V}$ in
\eqref{eq:latent_space_separation}, on the other hand, does not include any
vectors from $U$ and, as a result, does not have any information on the observed
data. Our approach to mapping of unseen vectors, therefore, is to locate a
latent vector $\bm{z}$ of an unseen vector within an open ball in $\hat{U}$
(i.e., $\bm{z}{\in}B_d(\bm{u}, 2\varepsilon), \exists \bm{u}{\in}U$) through the
transformation technique described in Section~\ref{sec:transformation}; in this
way, due to the continuity between the latent space and the output space,
$g_\theta(\bm{z})$ would be close to $g_\theta(\bm{u})$.
\begin{figure}[t!]
  \begin{center}
    \includegraphics[width=0.8\linewidth]{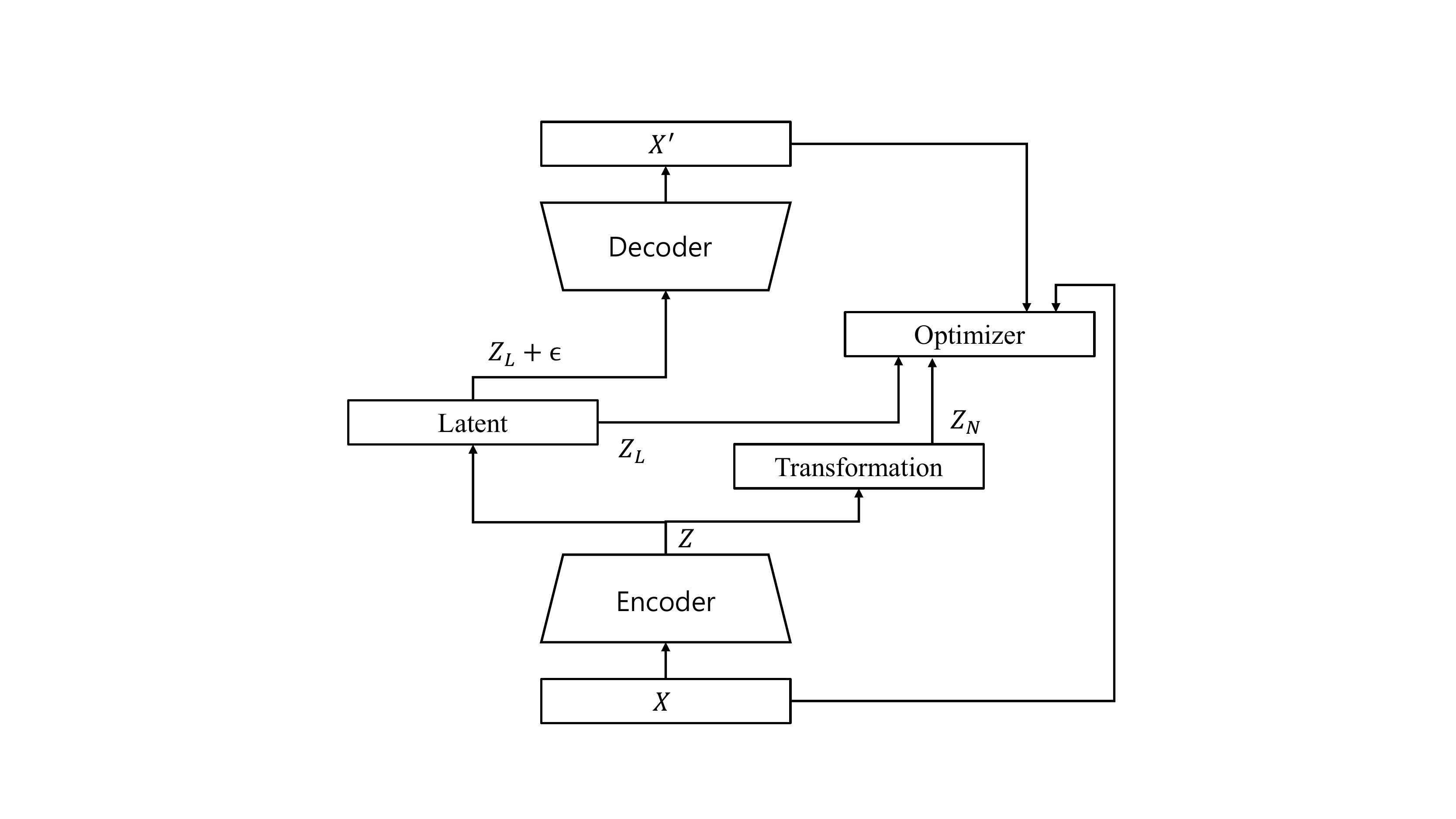}
  \end{center}
  \vspace{-20pt}
  \caption{Architecture of the Latent space Transformation Autoencoder. $Z$,
    $Z_N$ and $Z_L$ denote spaces of the outputs of the encoder, the
    transformation and the latent networks. $\epsilon$ is a set of noise which
    has the same size with the $Z_L$. Vectors in $Z$ go to the latent network
    and the transformation network. Vectors in $Z_L$ are learned to form as
    similar as possible with corresponding vectors in $Z_N$ and to recover
    vectors in $X'$ as similar as possible to corresponding vectors in $X$.}
  \label{fig:architecture}
\end{figure}

\subsection{Transformation}
\label{sec:transformation}
Note that $\hat{U}$ is not appropriate for an input space of a generative model,
because $\operatorname{diam}(\hat{U})$ is big and clusters of vectors in the
latent space are far away from each other in general, which makes it difficult
to interpolate. In this section, we define the transformation network and the
latent network in order to transform vectors in $Z$ into $Z_L$.

Through the latent network, $\operatorname{diam}(\hat{U})$ becomes smaller and
clusters of vectors in $Z$ get closer. The transformation network and the latent
network are located between the encoder and the decoder network. The outputs of
the encoder network go to the transformation network and the latent network. Let
$X \in X_{in}$ be a set of input vectors at an
iteration. Figure~\ref{fig:architecture} shows the architecture of the LTAE. In
the transformation network, each element of $\bm{z}$ in $Z$ is transformed by
the standard normalization\footnote{Note that we choose the standard
  normalization and the min-max normalization for a simple transformation
  case. There is no limitation of transformation methods. Any transformation
  technique that makes vectors close without changes of topological properties
  can be used.}
\begin{equation}
  (z_N)_i = \frac{z_i-\mu (Z_i)}{\sigma (Z_i)},
\end{equation}
where $\mu (\cdot)$ and $\sigma (\cdot)$ denote the mean and the standard
deviation, or by the min-max normalization,
\begin{equation}
  (z_N)_i = \frac{z_i-\min(Z_i)}{\max(Z_i)-\min(Z_i)},
\end{equation}
where $Z_i$ is a set of the $i$th element of all vectors in $Z$ and
$\bm{z}_N{=}((z_N)_i) \in Z_N$.

We want to train the parameters of the latent network to compute the
normalization process of the transformation network. Because both normalization
methods require subtract first and then division, we define two $m$-dimensional
row vectors $\bm{\alpha}$ and $\bm{\beta}$ of the latent network such that
$\bm{z}_L$ is calculated by
\begin{equation}
  \label{eq:transform}
  \bm{z}_L = \bm{\alpha}\bigodot(\bm{z}\bigoplus\bm{\beta}),
\end{equation}
where $\bigodot$ and $\bigoplus$ denote element-wise multiplication and
addition. Then, the $L_2$ loss between $Z_L$ and $Z_N$ is calculated so that
$\bm{\alpha}$ and $\bm{\beta}$ are learned to make $Z_L$ and $Z_N$ similar. Note
that we cannot use probability-based loss functions like cross-entropy for the
loss between $Z_L$ and $Z_N$ because a range of values of vectors are larger
than $[0,1]$.

Now, our goal is to train a neural network consisting of an encoder network $f$
and a decoder network $g$ with weights and biases $\phi$ and $\theta$, and the
transformation network output $Z_N$ and the latent network output $Z_L$ to
minimize the following loss function:

\begin{equation}
  \label{eq:goal}
  \frac{1}{N}\sum_{\bm{x} \in X_{in}} \bar{L}(\bm{x},g(\bm{z};\theta) ) +
  \|\bm{z_N} - \bm{z_L}\|_2 ,
\end{equation}
where $\bm{z}{=}\bm{u}{+}\bm{\epsilon}$ for a given $\varepsilon{>}0$ and
$\|\bm{\epsilon}\|{<}2\varepsilon$, $\bm{u}{=}f(\bm{x};\phi)$, and $\bar{L}$ is
either cross-entropy or $L_2$ loss.

\section{Analysis}
\label{sec:analysis}
The LTAE aims that clusters in the latent space get closer to one another and
thereby makes it easy to learn unseen vectors in the latent space so that any
vector in a specific subset of the latent space can have matched outputs. The
latent space and the transformed latent space share the same topological
properties because of the homeomorphism between the two spaces. The latent
network transforms vectors in the latent space into the transformed latent
space, where unseen vectors lie nearby observed vectors since
$\operatorname{diam}(Z_L)$ becomes small. All possible input vectors of the
decoder network during the generation process are sampled according to the
transformation used during the training process.

\subsection{Homeomorphism}
In topology, two homeomorphic spaces are considered to be topologically
equivalent. This means that, if topological space $X$ and $Y$ are homeomorphic,
all topological properties of $X$ (e.g., compactness, connectedness, or
Hausdorff) are preserved in $Y$.

Note that the equation (\ref{eq:transform}) can be rewritten as a function
$f{:}Z{\rightarrow}Z_L$: For $\bm{z}{\in}Z$ and $\bm{z}_{l}{\in}Z_{L}$, $f$ is
defined as $\bm{z}_l = f(\bm{z}) = \prod_{i=1}^m f_i(z_i)$,
where $f_i(z_i){=}\alpha_i{\cdot}(z_i{+}\beta_i)$ and $\prod$ denotes the
Cartesian product. Because $f$ is both continuous and bijection and has a
continuous inverse function (i.e., homeomorphism), $Z$ and $Z_L$ are
homeomorphic and topologically equivalent. With the transformation network,
$\bm{\alpha}$ is trained to get close with $1{/}\sigma(Z^{(i)})$ and
$1{/}\{\max(Z^{(i)}){-}\min(Z^{(i)})\}$, and $\bm{\beta}$ is trained to get
close with ${-}\mu(Z^{(i)})$ and ${-}\min(Z^{(i)})$ in the standard
normalization and the min-max normalization, respectively, while preserving the
topological properties of $Z$.

\subsection{Layer Transformation}
The layer transformation makes vectors in the latent space located within a
small and dense region. The method seems similar to batch normalization and
Layer normalization because the method calculates the standard normalization and
the min-max normalization \cite{ioffe2015batch,ba2016layer}. The main difference
between the layer transformation from batch normalization and the layer
normalization is that $\bm{\alpha}$ and $\bm{\beta}$ are learned to normalize
each output of the encoder network. The transformation network transforms
vectors in the latent space using statistical features of outputs of the encoder
network during the training and thereby the range of the latent vectors is
determined by the statistical features.
The main advantage of the layer transformation is that clusters of vectors get
close. As a result, distances between observed vectors in the latent space get
smaller and so it becomes easy to interpolate sparse spaces between all observed
vectors because if the vectors in the latent space are widely distributed during
the training process, many of them will not result in outputs close to those
corresponding to observed vectors in the input space during the generation
process. On the other hand, it would be easier for unseen vectors in the input
space to have outputs close to those corresponding to observed vectors in the
input space during the generation process, if the vectors corresponding to the
train dataset are close to one another in the latent
space.


\subsection{Similarity Measure of Sets of Images}
In evaluating the performance of generative models for image synthesis, two
major requirements, which are seemingly contradictive to each other, should be
taken into account: Generated images should have visual characteristics similar
to those of some training images, but, at the same time, differ from the
training images \cite{salimans2016improved}. In order to meet these
requirements, we propose the Hausdorff distance \cite{munkres1975topology} as a
metric capturing the similarity between two sets of images.

\begin{figure}[t]
  \begin{minipage}{0.3\linewidth}
    \centering
    \includegraphics[width=\linewidth]{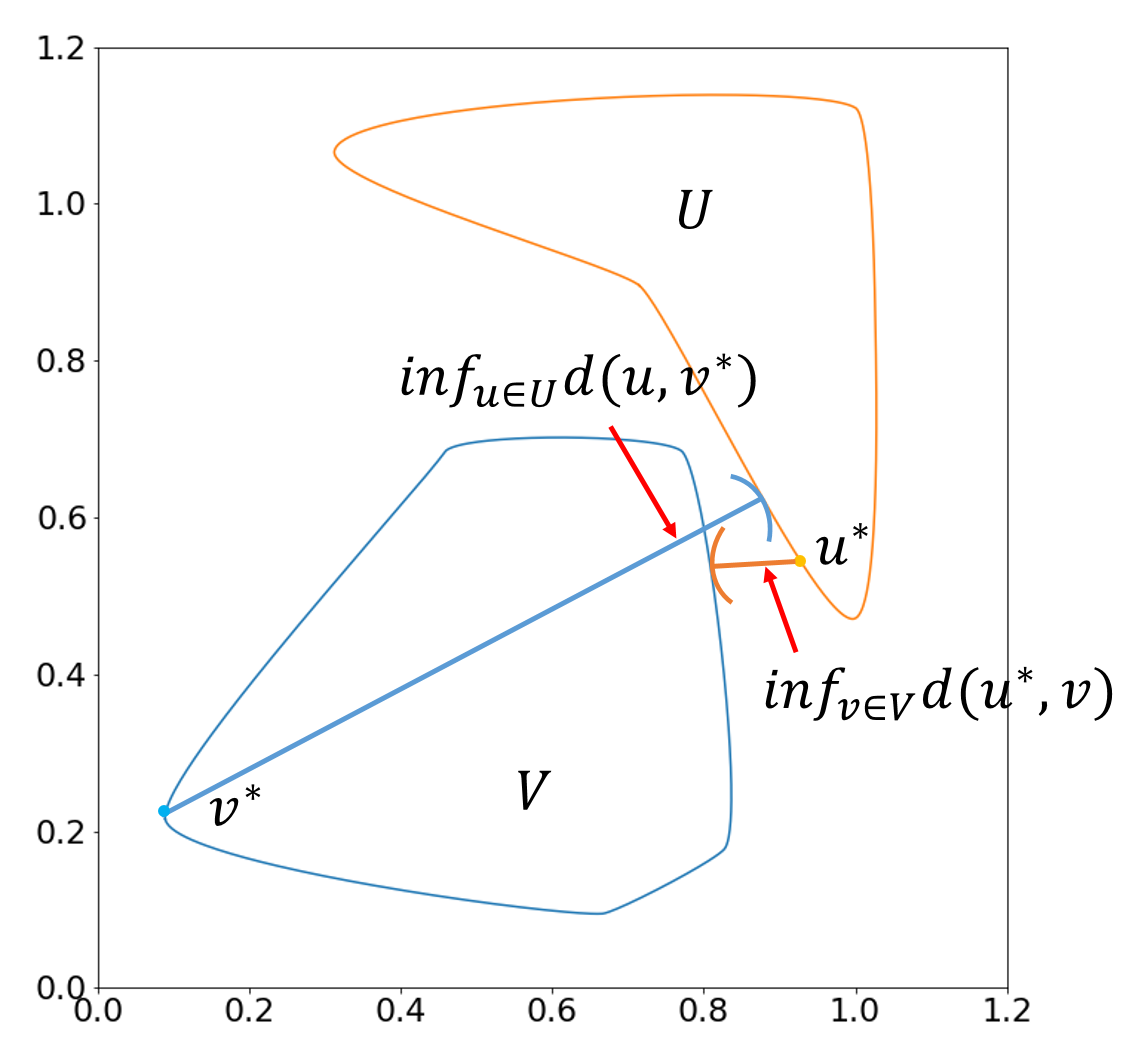}\\
    {\scriptsize (a)}
  \end{minipage}
  \hfill
  \begin{minipage}{0.3\linewidth}
    \centering
    \includegraphics[width=\linewidth]{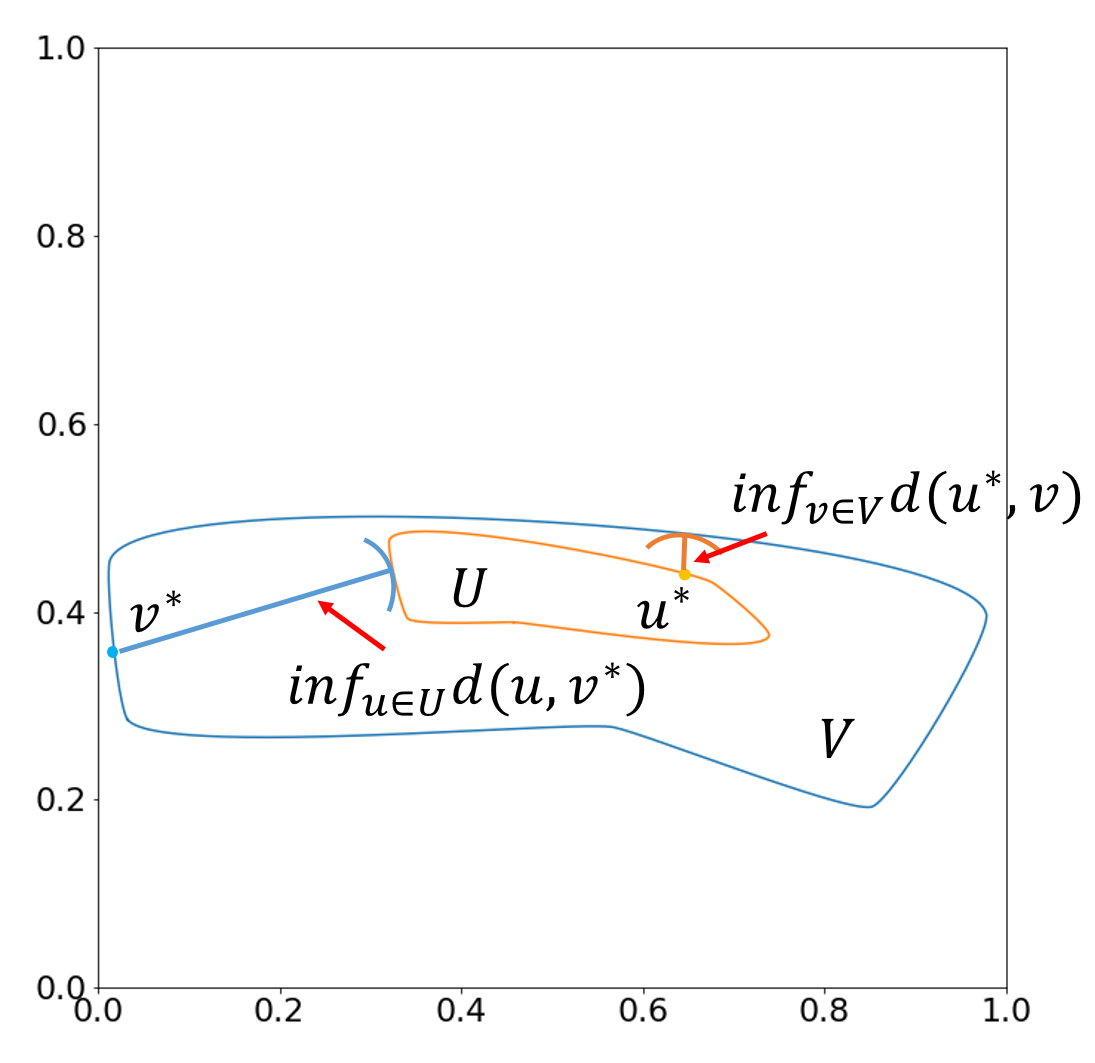}\\
    {\scriptsize (b)}
  \end{minipage}
  \hfill
  \begin{minipage}{0.3\linewidth}
    \centering
    \includegraphics[width=\linewidth]{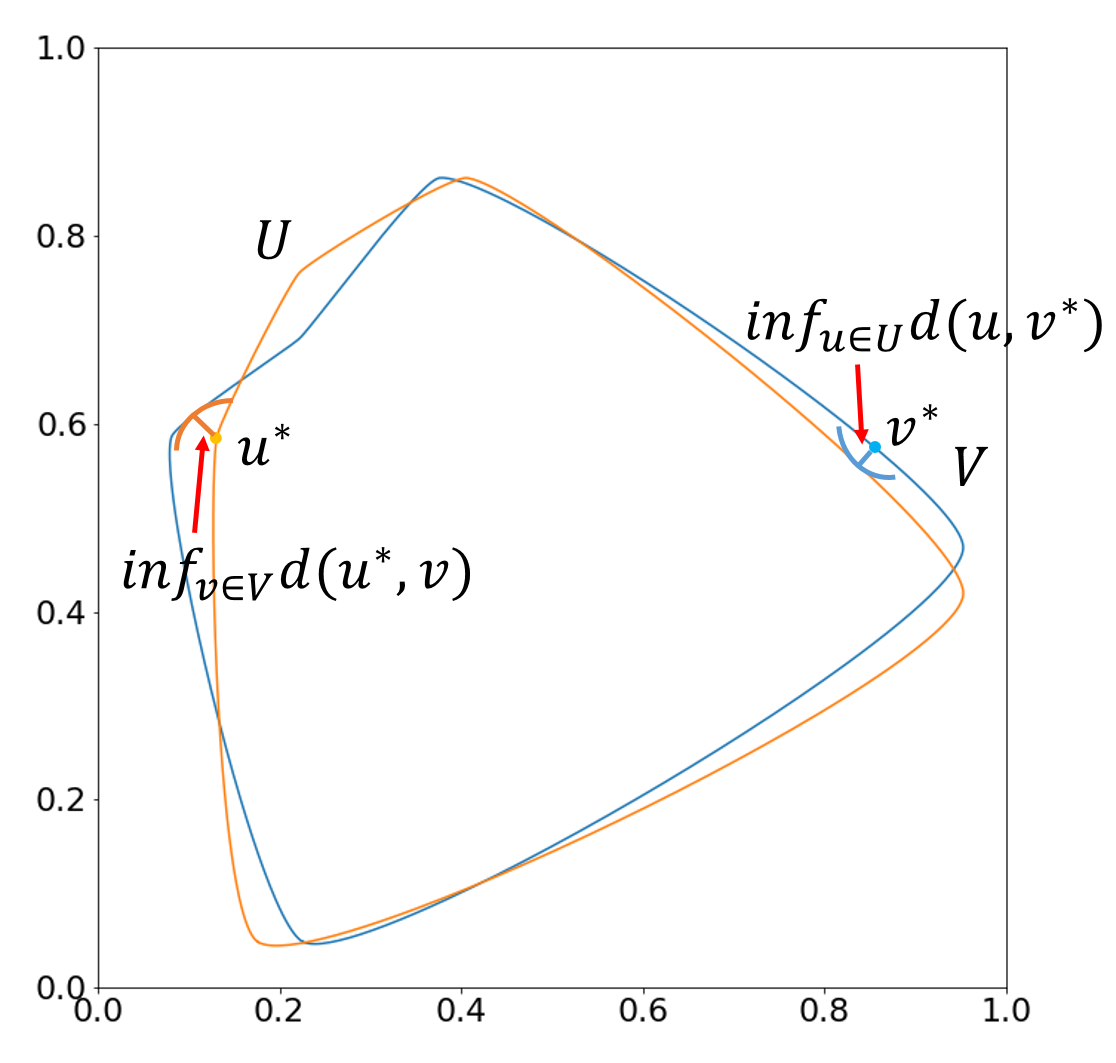}\\
    {\scriptsize (c)}
  \end{minipage}
  \caption{Examples of the Hausdorff distance between two non-empty subsets $U$
    and $V$ in space with the value of (a) 0.8672, (b) 0.3261 and (c) 0.0501.}
  \label{fig:hausdo}
\end{figure}

Let $U$ and $V$ be two different sets of images and $u$ and $v$ be individual
images belonging to $U$ and $V$, respectively.
Due to taking the maximum of both $\sup_{u\in U}\inf_{v\in V}d(u,v)$ and
$\sup_{v\in V}\inf_{u\in U}d(v,u)$ in the definition of the Hausdorff distance
given in \eqref{eq:hausdorff}, the difference of two sets of images can be
properly measured by taking into account the two major requirements. The
Hausdorff distances of three different types are illustrated in the
Figure~\ref{fig:hausdo}.

Note that, unlike the conventional use of the Hausdorff distance as a similarity
measure between two individual images/shapes (e.g.,
\cite{huttenlocher93:_compar_hausd}), we use it to measure the
\textit{similarity between two sets of images} (i.e., the set of training images
and that of processed images) to objectively evaluate the performance of
generative and denoising models.

\section{Experiment}
\label{sec:experiment}
We trained the LTAE model of images from the MNIST dataset\footnote{Available at
  http://www.cs.nyu.edu/~roweis/data.html}. The encoder and the decoder each has
two hidden layers with 500 hidden units for MNIST. The number of hidden units is
chosen based on prior autoencoder literature \cite{kingma2013auto}. A softplus
rectifier is used for two hidden layers in the encoder and the decoder. A linear
function is used for the output layer of the encoder and a sigmoid function is
used for the output layer of the decoder. We use Cyclical Learning Rates (CLR)
for with the base learning rate 0.001, the maximum learning rate 0.005, and step
size of 5500 \cite{smith2017cyclical} with batch size of 100 and. The weights are
initialized by Xavier initialization and the . The model is tested with different values
of $\varepsilon$ and latent space dimension.

In this paper, we use the standard normalization and the min-max normalization
transformation technique. $\varepsilon$ is added to variables at the transformed
latent space so that the LTAE learns unseen data around the input data set while
training, where $\varepsilon{\sim}\mathcal{N}(0, \sigma^2)$ or
$\varepsilon{\sim}\mathcal{U}(-\sigma, \sigma)$ according to the transformation
technique. We use LTAE-S-$\sigma$ and LTAE-M-$\sigma$ to denote the LTAE with
$\sigma$ by the standard normalization and the min-max normalization
transformation technique, respectively.


\begin{figure}[t]
  \begin{minipage}{0.32\linewidth}
    \centering
    \includegraphics[width=\linewidth]{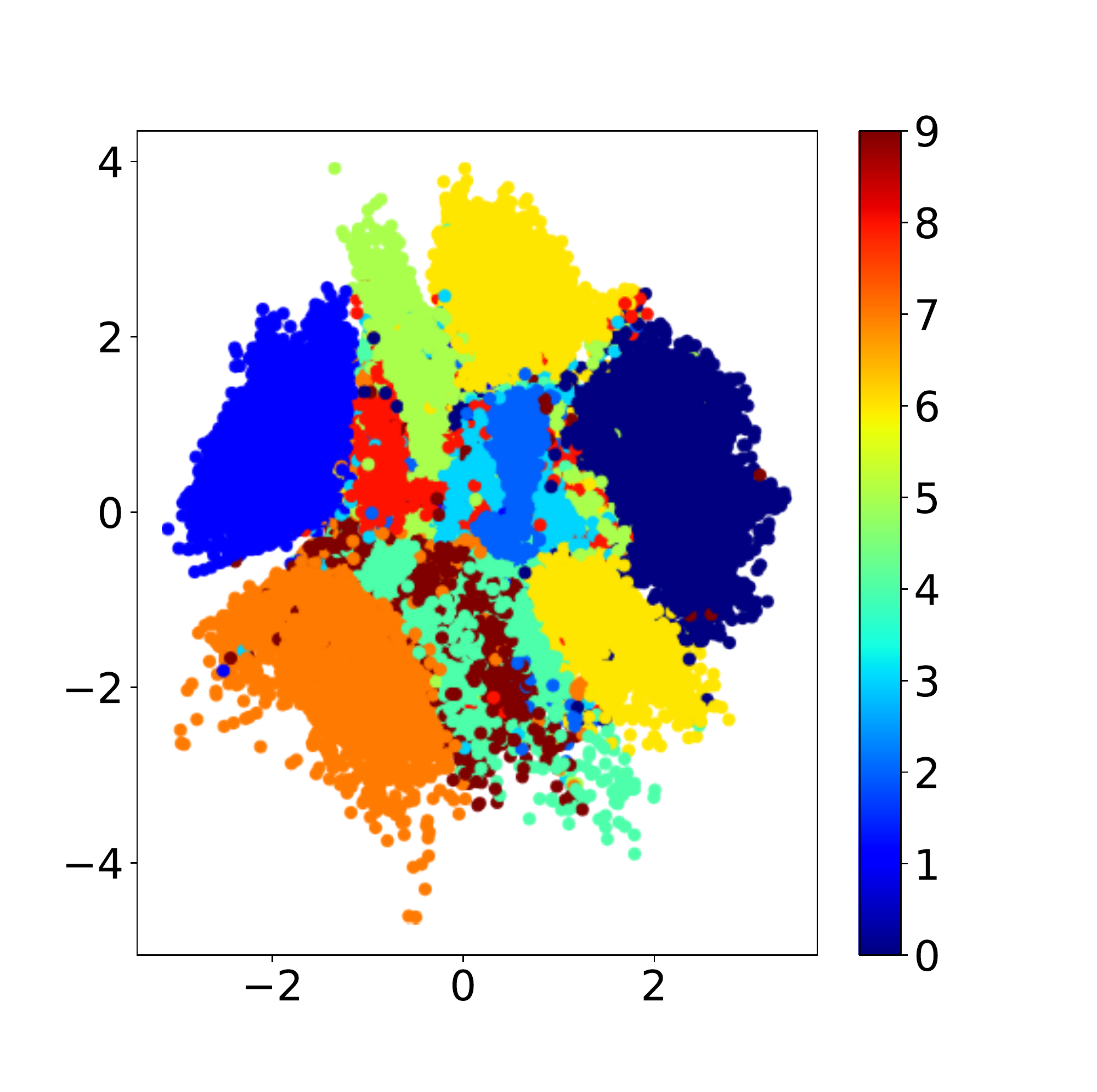}\\
    {\scriptsize (a)}
  \end{minipage}
  \hfill
  \begin{minipage}{0.32\linewidth}
    \centering
    \includegraphics[width=\linewidth]{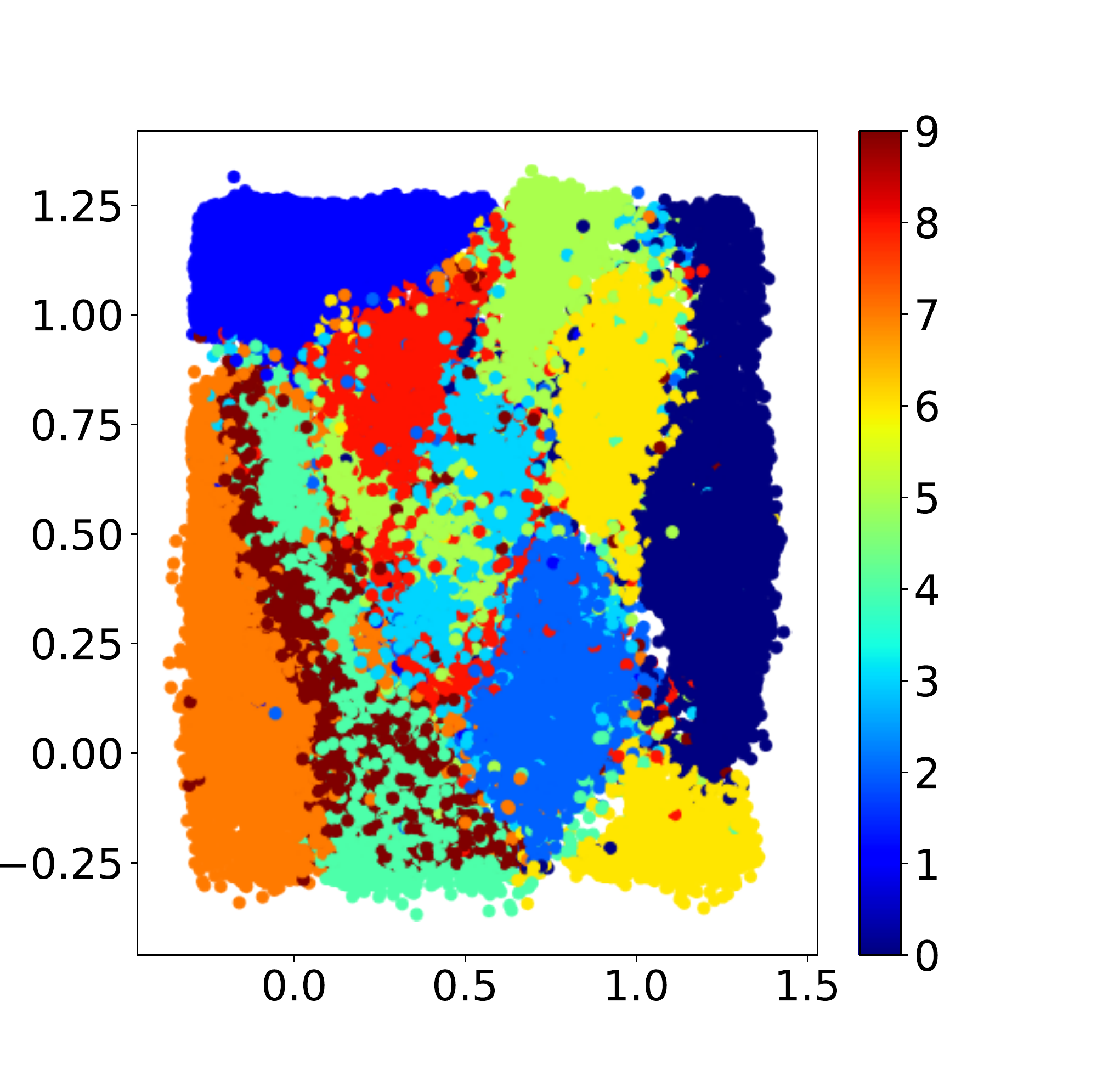}\\
    {\scriptsize (b)}
  \end{minipage}
  \hfill
  \begin{minipage}{0.32\linewidth}
    \centering
    \includegraphics[width=\linewidth]{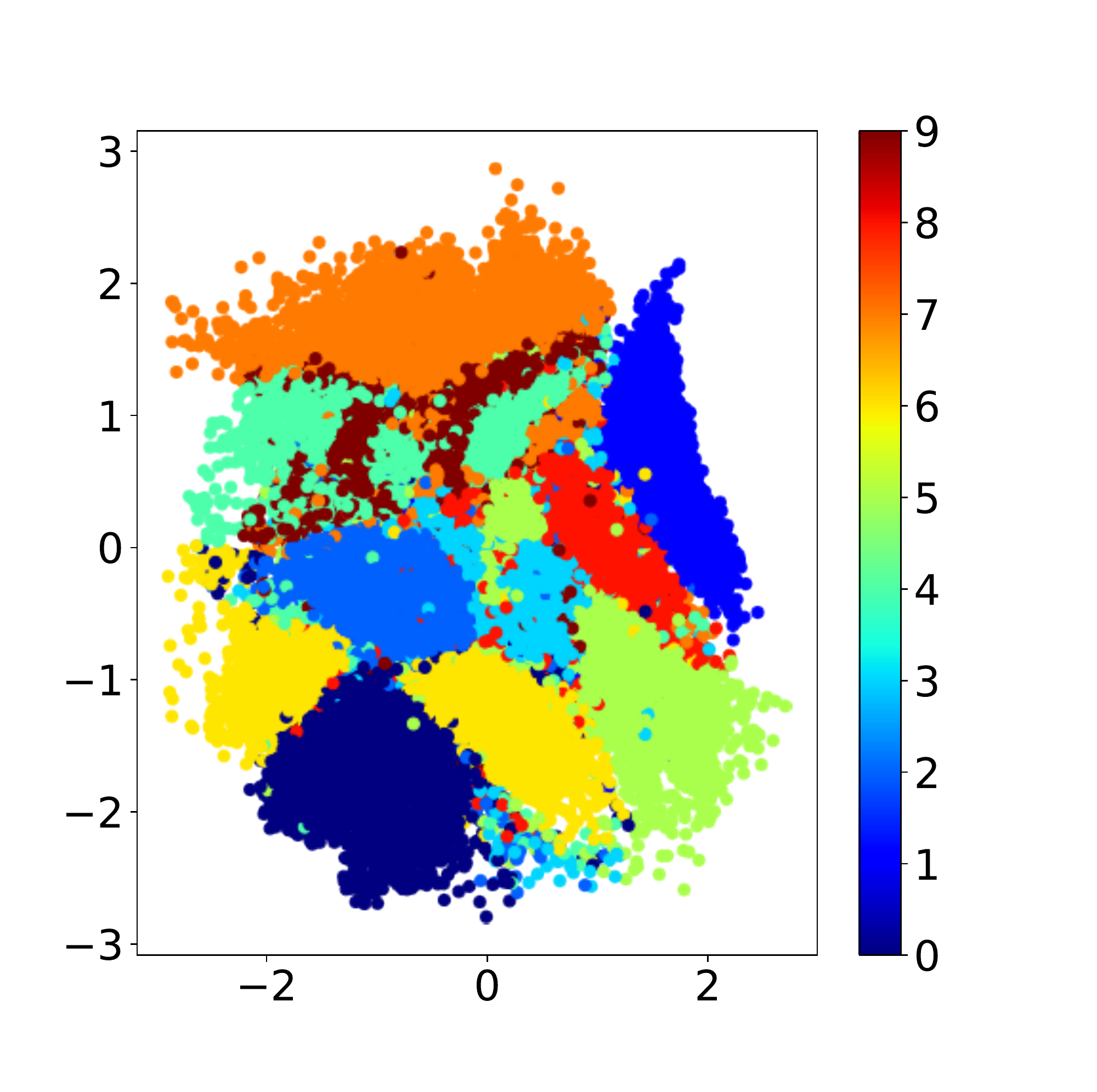}\\
    {\scriptsize (c)}
  \end{minipage}
  \caption{Transformed latent space of (a) 2d-VAE (b) 2d-LTAE-M-0.06, and (c)
    2d-LTAE-S-0.02. The range of the latent vectors varies depending on the
    transformation technique.}
  \label{fig:latent}
\end{figure}

\subsection{Generative Models}
\label{sec:generative}
In order to generate images, vectors are sampled with respect to the
transformation techniques. Vectors are sampled from
$\mathcal{N}(\bm{\mu}, \bm{\sigma}^2)$ in the LTAE-S, where $\bm{\mu}$ and
$\bm{\sigma}$ are the mean and the standard deviation of vectors in $Z_L$,
respectively. In the LTAE-M, vectors are sampled from
$\mathcal{U}(\bm{m}, \bm{M})$, where $\bm{m}$ and $\bm{M}$ are the minimum and
the maximum of the vectors in $Z_L$, respectively.

\subsection{DLTAE: Denoising and Generative Models}
The transformation and the latent networks can be located between any layers,
which makes it easy to combine LTAE with any AEs. We propose DLTAE by
introducing noise as same as the DAE
\cite{vincent2008extracting,vincent2010stacked}. In this paper, we take an
example of the simple DAE case whose noise is injected only at the input
space\footnote{It is not the limitation of the DLTAE. Any corruption process can
  be applied to the DLTAE.}. The architecture of the DLTAE is the same as the
LTAE and corruption data process is the same as the DAE: The corrupted input
vector by added noise, i.e., $\bm{x}{+}\bm{\hat{\epsilon}}$, is injected to the
encoder network. The output of the decoder network of
$\bm{x}{+}\bm{\hat{\epsilon}}$ is compared with the original vector,
$\bm{x}$. The loss function of the DLTAE is the same with the Equation
(\ref{eq:goal})
%
except $\bm{u}{=}f(\bm{x}{+}\bm{\hat{\epsilon}};\phi)$ and $\bar{L}$ is the
cross-entropy loss in our experiment
In fact, compared to the DAE, the reduction of the introduced noise in DLTAE
occurs at two different places, i.e., the encoder network related with the noise
occurs at input space and the decoder network in regard to the noise at latent
space. Due to the corruption of inputs and its same structure with the LTAE, the
DLTAE can be used as a denoising model and a generative model at the same time.

\begin{table}
  \caption{The Hausdorff distance between training images and generated images}
  \label{tbl:Gen_compar}
  \centering
  \begin{tabular}{l|lll}
    \toprule
    Compared with & VAE & DLTAE-M & DLTAE-S \\
    \midrule
    Hausdorff distance ($L_2$-norm)  & 9.3641 & 7.4631  & 8.7427    \\
    \midrule
    Hausdorff distance (cross-entropy)  & 6.2191 & 5.8758 & 5.6153\\
    \bottomrule
  \end{tabular}
\end{table}

\subsection{Comparison with VAE for Generative Models}
We take the VAE and calculate the Hausdorff distance by taking $L_2$-norm and
cross entropy as $d$ in the equation (\ref{eq:hausdorff}) between the training image set and
the generated image set for a comparison with the proposed model for a
generative model.
We train the VAE with the same number of hidden layers and units. The base
learning rate and the maximum learning rate are set 0.0008 and 0.002,
respectively, because the gradient decent diverges while training the VAE with
the same learning rate condition mentioned  in Section~\ref{sec:experiment}.
Transformed latent space of the LTAE-M-0.06, the LTAE-S-0.02, and the VAE with 2
dimensional latent space are shown in Figure~\ref{fig:latent}. The range of the
transformed latent space are determined according to the transformation
technique.

We take the DLTAE-M and DLTAE-S to compare with the VAE for a generation
performance. First, we compare three models with MNIST data set. 100 samples are
randomly picked up and then used to train three models with 2 dimensional latent
space. We change the step size for CLR to 10 and iterations to 40000 because the
number of the data set has been changed. Ten sets of 10000 images are generated
and the mean of the Hausdorff distance between training images and each set is
summarized in Table~\ref{tbl:Gen_compar}.

\begin{table}
  \caption{The Hausdorff distance of corrupted images and reconstructed images
    with respect to training images}
  \label{tbl:Recon_compar}
  \centering
  \begin{tabular}{l|llll}
    \toprule
    Compared with & Corrupted images & DAE & DLTAE-M & DLTAE-S \\
    \midrule
    Hausdorff distance ($L_2$-norm) & 14.8508 & 6.2902  & 4.7362  & 4.6981  \\
    \midrule
    Hausdorff distance (cross-entropy) & 5.7924 & 5.4895  & 5.3686 & 5.3655 \\    
    \bottomrule
  \end{tabular}
\end{table}
\subsection{Comparison with DAE for Denoising Models}
Even though the goal of the DAE is not reconstruction, we compare the
reconstruction of the DLTAE with the DAE to check its denoising performance. The
DAE is trained with the same number of hidden layers, units, and hyperparameters
for CLR. While training, Gaussian random noise of $\mathcal{N}(0,0.5^2)$ is
added to an original input image. The corrupted image is injected to the three
models. The sample outputs from the three models are shown in Appendix, and the
Hausdorff distances are summarized in Table~\ref{tbl:Recon_compar}, which
demonstrate that the reconstruction images by the DLTAE is more similar to the
original images by capturing salient features like the DAE.

\section{Conclusions}
\label{sec:conclusions}
We have proposed a novel framework for AE based on the homeomorphic
transformation of latent variables through new latent and transformation
networks installed between the encoder and the decoder networks of AE; unlike
the conventional VAEs based on the reparameterization trick with independent
Gaussian latent variables, the proposed framework allows more flexibility in
handling the latent space while maintaining the direct connection from inputs to
latent variables to outputs. We have investigated the effect of the
transformation in both learning generative models and denoising corrupted data.
The experimental results with the images from the MNIST dataset
show that the proposed framework could generate a model working as both a
generative model and a denoising model with much improved performance.

\bibliographystyle{plainnat}
\bibliography{IEEEabrv,neurips_2019}

\end{document}